# Exact Histogram Specification Optimized for Structural Similarity

Alireza Avanaki

userID at yahoo.com (my ID is the same as my last name)

**Abstract**

**An exact histogram specification (EHS) method modifies its input image to have a specified histogram. Applications of EHS include image (contrast) enhancement (e.g., by histogram equalization) and histogram watermarking. Performing EHS on an image, however, reduces its visual quality. Starting from the output of a generic EHS method, we maximize the structural similarity index (SSIM) between the original image (before EHS) and the result of EHS iteratively. Essential in this process is the computationally simple and accurate formula we derive for SSIM gradient. As it is based on gradient ascent, the proposed EHS always converges. Experimental results confirm that while obtaining the histogram exactly as specified, the proposed method invariably outperforms the existing methods in terms of visual quality of the result. The computational complexity of the proposed method is shown to be of the same order as that of the existing methods.**

## I. Introduction and Background

Exact histogram specification (EHS) refers to the problem of changing the input image so that the histogram of the resulted image matches a specified (a.k.a. target) histogram. EHS is an ill-posed problem as the number of images having the target histogram is very large (Section IV). Among these images, the ideal EHS method should find the one most similar to the original image. Because of its extensive utility in image processing (Section I.a), histogram specification received considerable attention [1-2, 7-8, 15-17]. Histogram equalization, a popular image enhancement operation, is a special case of histogram specification with a flat target histogram.

---

**Algorithm 1**: Prototype of classic and modern exact histogram specification (EHS) methods

Step 0: Let the target histogram be $H = \{h_0, h_1, ..., h_{L-1}\}$. $L$ is the number of possible intensity levels (e.g., 256 in 8-bit image). We assume that $\sum_{i=0}^{L-1} h_i = M$, in which $M$ is the number of pixels in the image. If $H$ does not satisfy this, scale it (and perhaps round some $h_i$) to realize this assumption.

Step 1: Sort the pixels of the original image by their intensity and auxiliary information in ascending order. (In classic method, aux. info = void)

Step 2: Starting from the first pixel on the sorted list, assign the first $h_0$ pixels a new intensity of 0. Continue by assigning the next $h_1$ pixels a new intensity of 1, and so on until all pixels are assigned their new intensities.

---

The state-of-the-art EHS methods [1, 2], with the generic structure given in Algorithm 1, are based on the concept of *strict ordering* of pixels [9]. Let us first describe strict ordering by the lack of it: in simple (classic) EHS, which is essentially a discrete version of probability density function (PDF) specification [9] (or [1]), the pixels are sorted by their intensities only (i.e., no auxiliary information is used in step 1 of Algorithm 1). When a group of pixels with the same original intensity has to be broken into two (or more) groups to fill in two (or more) different bins of the specified histogram, ambiguity arises: we do not know which pixel goes to which bin (i.e., which pixel gets which new intensity), as the only information used in sorting the pixels is their intensity which is constant within the group.

A modern EHS method establishes a strict order among the pixels, by using some auxiliary information in step 1 of Algorithm 1. This way, ideally, there will no ambiguity when a group of same-intensity pixels has to be broken among different bins of the target histogram: those with lower (higher) strict-order go to lower (higher) intensity bins. In [1], the auxiliary information required in step 1 is the local average intensities computed over six nested neighborhoods centered on each pixel. In [2], the auxiliary information comes from the wavelet coefficients corresponding to pixels in various subbands of the wavelet decomposition of the input image. In [1], the pixels with larger generalized intensities (i.e., original intensity + intensities over nested neighborhoods) may be assigned larger new intensities as compared to other pixels with the same original intensity. In [2], it is argued that the generalized intensity concept of [1] is oblivious to the structure of the image: consider two pixels of the original image with the same intensity, one from a smooth area and the other from an edge; instead of considering their generalized intensities, the method of [2] assigns their new intensities to keep the smooth area in the input image, smooth, and the edges of the input image, high contrast, in its output.

Another notable related work is [8]: the visual fidelity of EHS result is improved using projection onto convex sets (POCS) as a post-processing step. The problem with this approach is that the application of POCS changes the histogram of EHS output. Although the final result is of better quality, and the histogram of the result is a good approximation of the target histogram, the overall method is not an EHS. Nevertheless, [8] can be considered a good histogram specification method as compared to its elementary counterpart [23] which yields a crude approximation of the specified histogram.

Note that the EHS methods such as [7] that disregard the visual fidelity of the result in favor of processing speed are out of the scope of this work.

a) **Applications**

Image (contrast) enhancement ([12], [14], [17-19]) is a traditional usage of histogram equalization (i.e., EHS with flat target histogram). By applying histogram equalization, the same number of pixels is assigned to each and every possible intensity level. This results in a sharper image to a human observer as the otherwise indiscernible details (due to being in the very dark and very bright areas) become visible. Hence, histogram equalization is invaluable in displaying images that need to be analyzed and interpreted by human users (e.g., satellite and medical images).

Another application of EHS is histogram watermarking [11] (or [4], and [21-22]) in which the watermark is the specified histogram: this can be either the hidden message by itself, or the original histogram of the unmarked asset modulated by the hidden message. For example, it is shown that very robust (excluding histogram tampering from the attacks of course) watermarking can be achieved by making "holes" (i.e., empty bins) in the original histogram. In histogram watermarking, the embedding is carried out by EHS. In this application, the imperceptibility requirement of watermarking translates to the visual fidelity of the EHS output.

Some image coding (compression) methods perform better if EHS with a suitable target histogram is applied to their input first. In [20], for example, it is shown that a higher lossless compression ratio is achieved if the histogram of the input image is compacted before compression (i.e., less bins are used and no bins left empty in the midst). Histogram compaction may be reversed (with some loss, if there's no empty bin in the original histogram) after decompression to reproduce the original image (or a slightly distorted version, if histogram compaction is irreversible). A preprocessing by EHS can also boost the performance of image segmentation methods that assume their input image follows a certain distribution (e.g., mixture of Gaussians), by enforcing the assumed histogram.

**b) Motivation**

Our research is motivated by the attempts of [1] and [2] in retaining the structure of the original image as much as possible. We noticed that both methods, when it comes to maintaining visual fidelity, operate in an ad-hoc manner in the sense that they do not try to find the *image* most similar to the original among all images with the specified histogram. Instead, they try to find a better *strict ordering* among pixels. In this paper, we reformulate the problem of histogram specification as the image optimization problem just described: We measure the visual fidelity (i.e., the similarity of the EHS result with the original image) using structural similarity index (SSIM) [3] - a modern full-reference image quality metric that is strongly correlated with the perceptual image quality. We derive a closed-form formula for the gradient of SSIM index with respect to one of its input images and use it to adaptively increase SSIM of the output while keeping its histogram exactly as specified. This is possible thanks to computational simplicity of our formula for SSIM gradient that redirects the search for the highest SSIM in each iteration. Our experimental results show a considerable improvement of visual fidelity over the results reported by [1] and [2].

The idea of SSIM gradient ascent for quality improvement is not new: it was used in [3] to highlight SSIM's capability to find the best image among all images having the same mean square error with a reference image. However, to the best of our knowledge, the powerful method of SSIM gradient ascent is not used in any other application. That is perhaps because a closed-form and computationally simple formula for SSIM gradient was not developed, or because the usage of SSIM gradient ascent in [3] is not very inspiring: the approach as suggested in [3] cannot be employed in a real-world de-noising scenario where the original (noise-free) image that is required for computation of SSIM (and its gradient) is not available.

**c) Organization**

The rest of the paper is organized as follows. SSIM is defined in Section II and is decomposed into linear terms to facilitate calculation of its gradient derived in Section III. The proposed method is described in Section IV which also includes complexity and convergence analyses of the proposed method. Our experimental results are compared to those of the existing methods in Section V. Section VI concludes the paper with a summary of contributions and a discussion.

## II. Structural Similarity Index

The SSIM (a.k.a. Wang-Bovik) index [3] is defined by (1). If the input images are identical, the index is 1; and if they are uncorrelated the index is very small. If one of the input images is considered the reference, the index gives the quality of the other image as compared to the reference.

$$\text{SSIM}(x, y) = \frac{N(x, y)}{D(x, y)} = \frac{(2\mu_x \mu_y + C_1)(2\sigma_{xy} + C_2)}{(\mu_x^2 + \mu_y^2 + C_1)(\sigma_x^2 + \sigma_y^2 + C_2)}, \tag{1}$$

To capture the (dis-)similarity of the images better, the index is computed on windows sliding over the two images. Then all the resulted indexes are averaged to give one index. In that case, x and y represent windows (located at the same places) of the two input images. $\mu_x$ and $\mu_y$ are the average intensities of pixels in x and y, with standard deviations $\sigma_x$ and $\sigma_y$. $\sigma_{xy} = \frac{1}{M} \sum_{\forall i,j} (x(i,j) - \mu_x)(y(i,j) - \mu_y)$, in which $M$ is the number of pixels in x and $x(i,j)$ is the intensity of pixel $(i,j)$ and summation is performed over all pixels. $C_1$ and $C_2$ are small positive constants keeping the denominator non-zero. $N(x,y)$ and $D(x,y)$ denote the expressions in numerator and denominator respectively that are used later.

Although (1) gives the SSIM index between two windows, it can be also considered as the formula for the SSIM index map ($\text{SSIM}_{\text{map}}$), using element-wise addition and multiplication, with the parameters defined in (2). At each point, $\text{SSIM}_{\text{map}}$ is an indication of the local similarity of the input images.

$$\mu_y = w * y, \quad \sigma_y^2 = w * y^2 - \mu_y^2, \quad \sigma_{xy} = w * (xy) - \mu_x \mu_y \tag{2}$$

in which $w$ is a symmetric low-pass kernel. In the SSIM implementation that Wang provided online [10], $w$ is an 11x11 normalized Gaussian kernel. '$*$' denotes convolution. $\sigma_x^2$ and $\mu_x$ are defined similarly. Based on $\text{SSIM}_{\text{map}}$, the SSIM index between two images is defined as

$$\mathrm{SSIM}(x, y) = \frac{1}{M} \sum_{\forall i,j} \mathrm{SSIM}_{\mathrm{map}}(x, y; i, j) \tag{3}$$

Here, $M$ is the number of pixels in either of the input images, and $\mathrm{SSIM}_{\mathrm{map}}(x, y; i, j)$ is the value of SSIM index for windows centered at $(i, j)$ from input images x, y.

### III. Calculation of SSIM gradient

Based on our formulation of SSIM index given above, we first compute $\dfrac{d}{d\,Y(a,b)}\mathrm{SSIM}$ (arguments are dropped for convenience) and then $\nabla_Y \mathrm{SSIM}(I, Y)$.

$$\frac{d}{d\,Y(a,b)}\mathrm{SSIM} = \frac{1}{M} \sum_{\forall i,j} \frac{d}{d\,Y(a,b)}\mathrm{SSIM}_{\mathrm{map}} \tag{4}$$

in which

$$\frac{d}{d\,Y(a,b)}\mathrm{SSIM}_{\mathrm{map}} = \frac{\partial \mu_Y}{\partial Y(a,b)} \frac{\partial \mathrm{SSIM}_{\mathrm{map}}}{\partial \mu_Y} + \frac{\partial \sigma_{IY}}{\partial Y(a,b)} \frac{\partial \mathrm{SSIM}_{\mathrm{map}}}{\partial \sigma_{IY}} + \frac{\partial \sigma_Y^2}{\partial Y(a,b)} \frac{\partial \mathrm{SSIM}_{\mathrm{map}}}{\partial \sigma_Y^2}. \tag{5}$$

By calculation of partial derivates of the parameters defined in (2) w.r.t. $Y(a,b)$, we have:

$$\frac{\partial \mu_Y}{\partial Y(a,b)} = \frac{\partial(w*Y)}{\partial Y(a,b)} = w(i,j) * \frac{\partial Y}{\partial Y(a,b)} = w(i,j) * \delta(i-a, j-b) = w(i-a, j-b) \tag{6.a}$$

$$\frac{\partial \sigma_{IY}}{\partial Y(a,b)} = w(i-a, j-b)\big(I(a,b) - \mu_I\big) \tag{6.b}$$

$$\frac{\partial \sigma_Y^2}{\partial Y(a,b)} = 2w(i-a, j-b)\big(Y(a,b) - \mu_Y\big) \tag{6.c}$$

The detailed calculation is only shown in (6.a), where the second equivalence holds because convolution and partial derivative (both linear operators) can be interchanged; the rest are derived similarly. $\delta(.,.)$ denotes discrete 2-D Dirac's delta function defined as:

$$\delta(i-a, j-b) = \begin{cases} 1 & i=a, j=b \\ 0 & \text{otherwise} \end{cases}.$$

By substituting the partial derivates into (5) and collecting $w(i-a, j-b)$, the summation in (4) turns into a weighted sum of three convolutions:

$$M \nabla_Y \text{SSIM}(I,Y) = w * M_1 + \left( w * \frac{\partial \text{SSIM}_{map}}{\partial \sigma_{IY}} \right) I + 2 \left( w * \frac{\partial \text{SSIM}_{map}}{\partial \sigma_Y^2} \right) Y, \quad \text{and}$$

$$M_1 = \frac{\partial \text{SSIM}_{map}}{\partial \mu_Y} - \mu_I \frac{\partial \text{SSIM}_{map}}{\partial \sigma_{YI}} - 2\mu_Y \frac{\partial \text{SSIM}_{map}}{\partial \sigma_Y^2}. \qquad (7)$$

Using (7), we can calculate $\nabla_Y \text{SSIM}(I,Y)$ for all pixels with just three convolutions (SSIM needs five) and some element-wise multiplications and additions. The partial derivates required in (7) and the simplified auxiliary variable $M_1$ are given below.

$$M_1 = \frac{2\mu_I(2\sigma_{IY} + C_2 - 2\mu_I\mu_Y - C_1) - 2\mu_Y(\sigma_I^2 + \sigma_Y^2 + C_2 - \mu_I^2 - \mu_Y^2 - C_1)\text{SSIM}_{map}}{D(I,Y)},$$

$$\frac{\partial \text{SSIM}_{map}}{\partial \sigma_{YI}} = \frac{2(C_1 + 2\mu_I\mu_Y)}{D(I,Y)}, \text{ and } \frac{\partial \text{SSIM}_{map}}{\partial \sigma_Y^2} = \frac{-\text{SSIM}_{map}}{\sigma_I^2 + \sigma_Y^2 + C_2}, \qquad (8)$$

in which $D(I,Y)$ is defined in (1). To gain further computational savings, one can compute $\text{SSIM}(I,Y)$ and $\nabla_Y \text{SSIM}(I,Y)$ in one procedure as the same intermediate variables are required in computing both: the runtime of our code (provided in [5]) that computes both $\text{SSIM}(I,Y)$ and $\nabla_Y \text{SSIM}(I,Y)$ is only 70% more than that of the code provided in [10] (which computes SSIM index only), for 512 x 512 and 1024 x 1024 images.

### IV. Proposed Method

Let *I* be a *M*-pixel gray-scale image with *L* possible levels of intensity for each pixel and an absolute histogram $H = \{h_0, h_1, ..., h_{L-1}\}$. The number of images that fit these descriptions is given by

$$\binom{M}{h_0} + \binom{M-h_0}{h_1} + \binom{M-h_0-h_1}{h_2} + \ldots + \binom{h_{L-1}+h_{L-2}}{h_{L-2}} + 1, \text{ in which } \binom{k}{l} = \frac{k!}{l!(k-l)!}$$

This is a very large number even for small images (e.g., about $4.5 \times 10^{15}$, for a 64x64 8-bit image). Thus, performing the ideal EHS (i.e., searching among all these images for the one that is most similar to the input image) can be quite difficult. In such a large solution space, not only exhaustive search is not an option; even heuristic random search techniques such as genetic algorithm may not converge to a global optimum in a reasonable time. That is perhaps why the existing methods adhere to strict ordering of pixels rather than searching this large solution space.

---

**Algorithm 2**: The proposed method (SSIM gradient ascent in the subspace of images with histogram $H$)

Step 0: Let $I$ show the original input image. Set $X = I$.

Step 1: Apply an EHS method (e.g. [1]) to $X$ to generate image $Y$ with given histogram $H$.

Step 2: Compute $\nabla_Y \text{SSIM}(I,Y)$ and $\text{SSIM}(I,Y)$.

Step 3: If convergence is reached, then break.

Step 4: Set $X = Y + \mu M \nabla_Y \text{SSIM}(I,Y)$ and go to Step 1.

---

Our proposed method is summarized in Algorithm 2, in which $\mu$ is a positive constant (step size), $\nabla_Y$ denotes gradient with respect to image $Y$ (i.e., $\nabla_Y(i,j) = \frac{\partial}{\partial Y(i,j)}$), and $M$ is the number of pixels in $I$. $\text{SSIM}(I,Y)$ is a measure of perceptual similarity of $I$ and $Y$ (the higher SSIM, the more similar are its input images). Our method is a simple hill climbing: the updated $X$ in step 4 is more similar to $I$ as we changed each pixel of $Y$ so that the SSIM is increased (i.e., we move in the direction of SSIM gradient). As it is inversely proportional to $M$, SSIM gradient (given by (7)) is scaled by $M$ so that $\mu$ becomes independent of image size. The histogram of the visually improved image $X$, however, may be a bit different than $H$; this is fixed in step 1 of the next iteration. The convergence criterion in step 3 may be either (or a combination) of the following: (i) the output quality is good enough (i.e., $\text{SSIM}(I,Y) >$ threshold). (ii) The number of iterations has reached a limit (good to control the overall complexity; see Section V.a).

Note that our method is based on the premise that the most similar image with the specified histogram is not very different from the result of an existing EHS method (e.g., classic, [1] or [2]). In other words, the correction made to $Y$ in step 4 to make it more similar to $I$ is not undone by performing EHS in step 1 in the next iteration. Our experimental results (Section V) verify that this is a reasonable assumption and that the good performance of Algorithm 2 does not depend on the type of EHS (classic or based on strict ordering) used in step 1.

**a) Complexity Analysis**

In addition to the computations required for EHS in step 1, in each iteration, we need some $O(M)$ computations in steps 2 and 4 ($M$ is the number of pixels). The simplest EHS method involves sorting that requires $O(M \log M)$ computations. Thus, the complexity of each iteration of Algorithm 2 is $O(M \log M)$. The overall complexity of Algorithm 2 is also $O(M \log M)$, assuming that the number of iteration to reach convergence does not depend on $M$. This assumption is validated experimentally in Section V, where also the runtime of the proposed method is compared to the existing methods for various image sizes.

**b) Convergence Analysis**

The main loop of Algorithm 2 can be summarized as

$$Y(n) = \text{EHS}(X(n), H) \qquad (9.\text{a})$$

$$X(n+1) = Y(n) + \mu M \nabla_Y \text{SSIM}(I, Y(n)) \qquad (9.\text{b})$$

in which EHS(.,.) represents exact histogram specification process, and $n$ is the iteration number. By substitution of (9.b) into (9.a), we have

$$Y(n+1) = \text{EHS}(Y(n) + \mu \, M \, \nabla_Y \, \text{SSIM}(I, Y(n)), H) \qquad (10)$$

Analyzing the convergence behavior of the dynamic process described by (10) is cumbersome since EHS is not differentiable with respect to its first argument. As we observe in Section V, the final result (i.e., $Y(n)$ for large $n$) depends on $\mu$. That is because although in the space of all images $\text{SSIM}(I,Y)$ is a smooth function of $Y$ with a single distinct maximum at $Y = I$, it is *not* smooth over the irregularly-shaped solution subspace (i.e., all images with histogram $H$). Therefore, in this subspace, there are several local maxima surrounding the global maximum of SSIM, to find which we have to try various values of $\mu$. In other words, the value of $\mu$ determines which local maximum the process of (10) converges to. That is when $\Delta Y(n) = \mu \, M \, \nabla_Y \, \text{SSIM}(I, Y(n))$ becomes smaller than the "dead-zone" of EHS input, thus the EHS output does not change. Fortunately, as it is shown in Section V, any reasonable choice of $\mu$ gives a very good suboptimal solution in a few iterations.

### IV.b.1. An upper bound on optimal step size

In the following, we develop an empirical method to find a good value of $\mu$ so that Algorithm 2 converges to a high SSIM quickly. The experiments that we based this method on are reported in Section V.c.

An estimate of SSIM increase due to the step 4 of Algorithm 2 at $n^{\text{th}}$ iteration is given by

$$\Delta\text{SSIM}(n) = \mu \, M \sum_{\forall i,j} (\nabla_Y \, \text{SSIM}(I, Y(n)))^2, \qquad (11)$$

in which power of 2 is performed element-wise. Note that this is just an estimate of $\Delta\text{SSIM}$ as $X$ should undergo EHS (in step 1) to make $Y(n+1)$ and because (11) is based on a first-order approximation.

A typical behavior of measured SSIM($n$) is depicted in Figure 1 (bottom-right). Hence, $\Delta\text{SSIM}(n)$ is well modeled by $\mu \, pq^n$. Therefore, after a large number of iteration we have:

$$\text{SSIM}_{\text{final}} = \text{SSIM}_{\text{init}} + \frac{p\mu M}{1-q} \qquad (12)$$

in which, $p = \sum_{\forall i,j} (\nabla_Y \text{SSIM}(I,Y(1)))^2$, $q = \frac{\Delta \text{SSIM}(2)}{\Delta \text{SSIM}(1)}$, and $\text{SSIM}_{\text{init}}$ is the value of SSIM index for the result of the first iteration. These parameters can be computed using the results of a three-iteration experimental run of Algorithm 2, for a typical value of $\mu$ (say 67). By substitution of the measured parameters in (12), approximating $\text{SSIM}_{\text{final}}$ by 1, and solving for $\mu$, we get an approximate upper bound on step size:

$$\mu_0 = \frac{1-q}{pM}(1 - \text{SSIM}_{\text{init}}) \qquad (13)$$

We observe that SSIM of the result starting from the $5^{\text{th}}$ iteration, is a convex function of $\mu$ in proximity of $\mu_0$ given by (13) (see Section V.c and Figure 2). Therefore, one has to search for $\mu$ in the vicinity of $\mu_0$ that gives the highest SSIM in five iterations and continue gradient ascent to reach the optimum solution (the image with the highest SSIM). In our experiments, the best $\mu$ is found within $[0.1\mu_0, \mu_0]$.

## V. Experimental Results

### a) Classic EHS suffices in step 1 of Algorithm 2

This experiment compares use of strict-ordering EHS of [1] and the classic EHS in step 1 of Algorithm 2. The input image is *cameraman* and the target histogram is that of *rice* (both are 8-bit 256 x 256), with $\mu = 67$ (see Sections IV.b.1 and V.c on choosing a good value for $\mu$). The results are shown in Figure 1, and the details (for one more test image and two more target histograms) are reported in Table 1. Superb visual quality of the outputs of Algorithm 2 is observed in Figure 1 (last column) and Table 1 (bottom row in each cell) as compared to the classic EHS and Coltuc's EHS [1] (top row in each cell). Since no significant difference between the SSIM indexes of the outputs (when compared to the original image) of the two variants of Algorithm 2 is observed, we use the classic EHS which is considerably computationally lighter (55% faster for the experiments reported in Table 1) in step 1 of Algorithm 2 in the rest of our experiments. This way, on a 1.66 GHz PC with 512 MB of RAM, each iteration of Matlab™ code needs 0.1227 sec CPU time, for a 256 x 256 image (see Table 2 for CPU time for other sizes). Typically, 90% of the total increase in SSIM occurs within 10-12 iterations (i.e., about 1.5 sec of CPU time; see Section V.c and Figure 3). Convergence to the final SSIM value may require 150-180 iterations (i.e., 22 sec of CPU time).

**Table 1.** Performance of Algorithm 2 (classic EHS left, Coltuc's [1] right) in terms of SSIM index (%) of the output at the first (top) and 180$^{th}$ (bottom) iterations, for three different target histograms (EHS with uniform target histogram = exact histogram equalization). $\mu = 67$ is used in all cases. It is observed that although Coltuc's EHS performs a bit better at first, it gives no significant improvement in the long run.

| Test image→ Target histogram↓ | *cameraman* | *lena* |
|---|---|---|
| *rice*'s | 70.81 v. 71.60 | 83.18 v. 83.54 |
|  | 83.89 v. 83.89 | 88.30 v. 88.32 |
| Linear | 77.68 v. 78.43 | 64.31 v. 64.58 |
|  | 82.33 v. 82.34 | 69.24 v. 69.22 |
| Uniform | 81.63 v. 83.27 | 77.03 v. 77.64 |
|  | 92.69 v. 92.72 | 84.83 v. 84.83 |

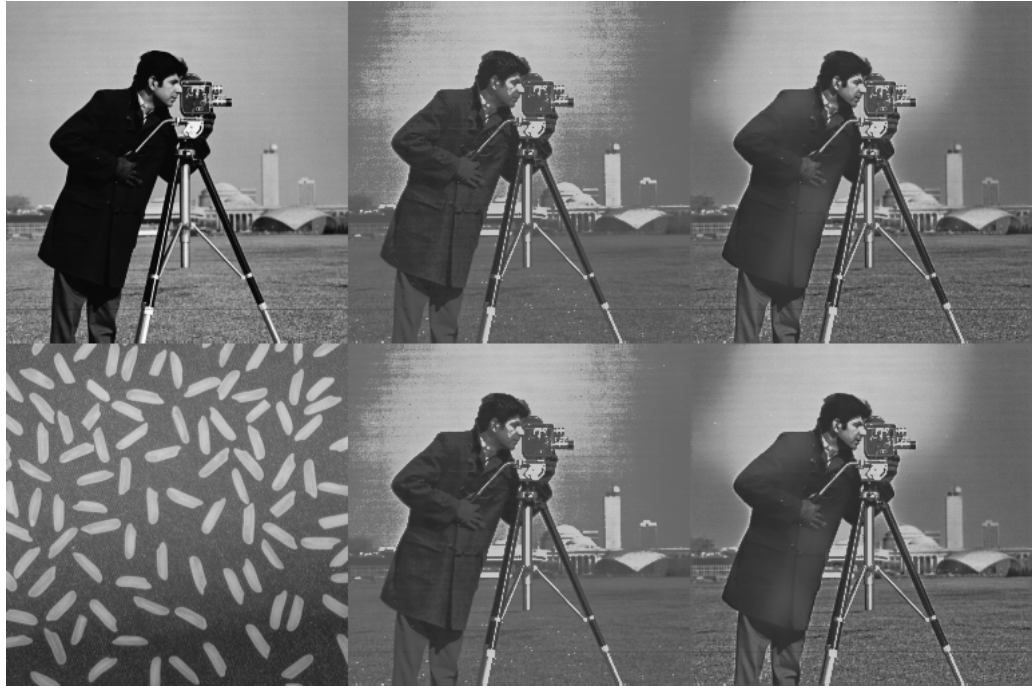

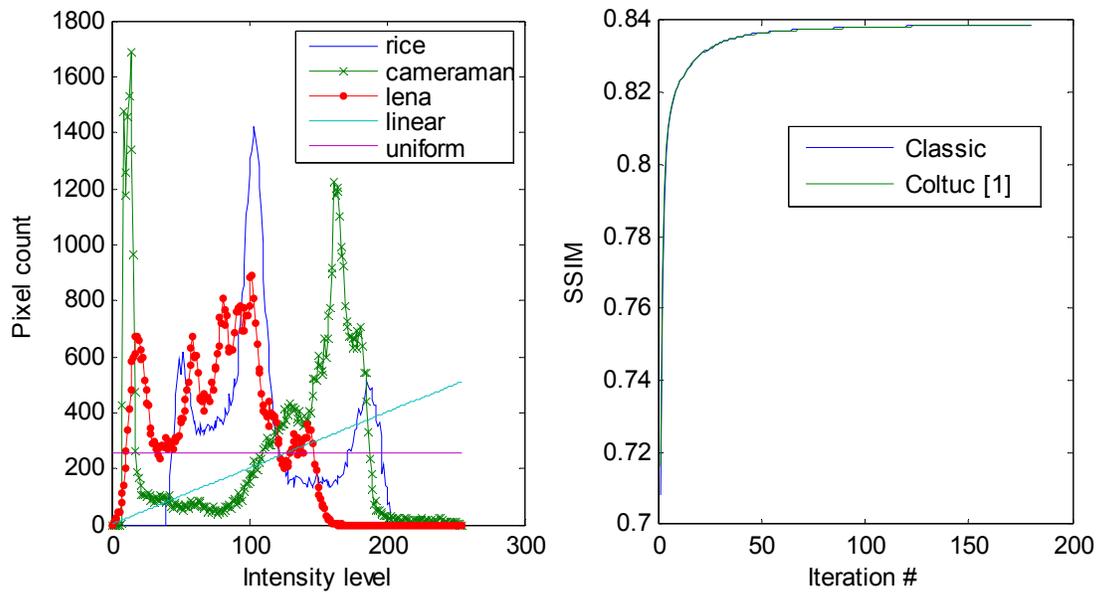

**Figure 1. Images** (from left to right, top to bottom): Original *cameraman* and *rice*, outputs of classic and Coltuc's EHS [1] methods with target histogram of *rice*, and outputs of Algorithm 2 (after 180 iterations) with classic and Coltuc's EHS used in step 1. It is observed that the great quality of the results (last column) does not depend on the choice of EHS used in Algorithm 2. **Graphs** (from left to right): Original and target histograms involved in the experiment, and SSIM index of the outputs of each iteration of Algorithm 2 run with classic and Coltuc's method for EHS in step 1.

**b) Application to histogram watermarking**

As mentioned earlier, a quality-optimized EHS method can be used to make histogram watermarking less visible. As an example, consider the watermark is the histogram of *rice*, to be embedded in the host asset *cameraman*. Figure 1 gives the watermarked images when the classic (top middle), Coltuc [1] (bottom middle), and the proposed method (right) EHS methods are used. It is observed that watermark embedding by our EHS is superior in preserving the visual fidelity of the watermarked image to the unmarked host.

**c) Choosing a good step size**

In this experiment, we study the dependence of the SSIM index of the result of each iteration on the iteration number and the value of $\mu$, for input images and target histograms used in the experiment of Section V.a. The results are shown in Figure 2. To achieve the best result (i.e., highest SSIM) or very close to the best, we observe, one can use the value of $\mu$ that maximizes SSIM within the first five iterations. It is also observed that, using any value of $\mu$, we can achieve some (considerable, in most cases) improvement over classic EHS (i.e., the result of $n = 1$). Finally, note that in the cases that the target histogram is considerably different (i.e., EHS largely affects many pixels; such as linear target histogram in our experiments) the result quality heavily depends on $\mu$, and the method requires more iterations to converge (SSIMs of $30^{th}$ and $60^{th}$ iterations are further apart as compared to the other cases).

To show the progressive improvement of visual quality, in Figure 3, the outputs for iterations 2 (i.e., the first improvement to classic EHS), 12, and 60 are shown for the cases that the highest and the lowest overall SSIM improvement are achieved in this experiment: *cameraman* with target histogram of *rice* (total relative improvement of 18.2%) and *cameramen* with linear target histogram (total relative improvement of 5.5%), respectively. Even in the latter case, the improvement is visible in the circled area. One argument that may be raised here is that the result of $60^{th}$ iteration for *cameraman* with linear target histogram is too "smoothed out" as compared to that of $12^{th}$ iteration. Note that this cannot be a fault with the proposed method, as it is trying to maximize the SSIM between the EHS output and the original image (0.8074 for $12^{th}$ and 0.8199 for $60^{th}$ iterations). Neither this can be considered SSIM's fault, as the structure of the overcoat is not very visible in the original image either (due to being very dark).

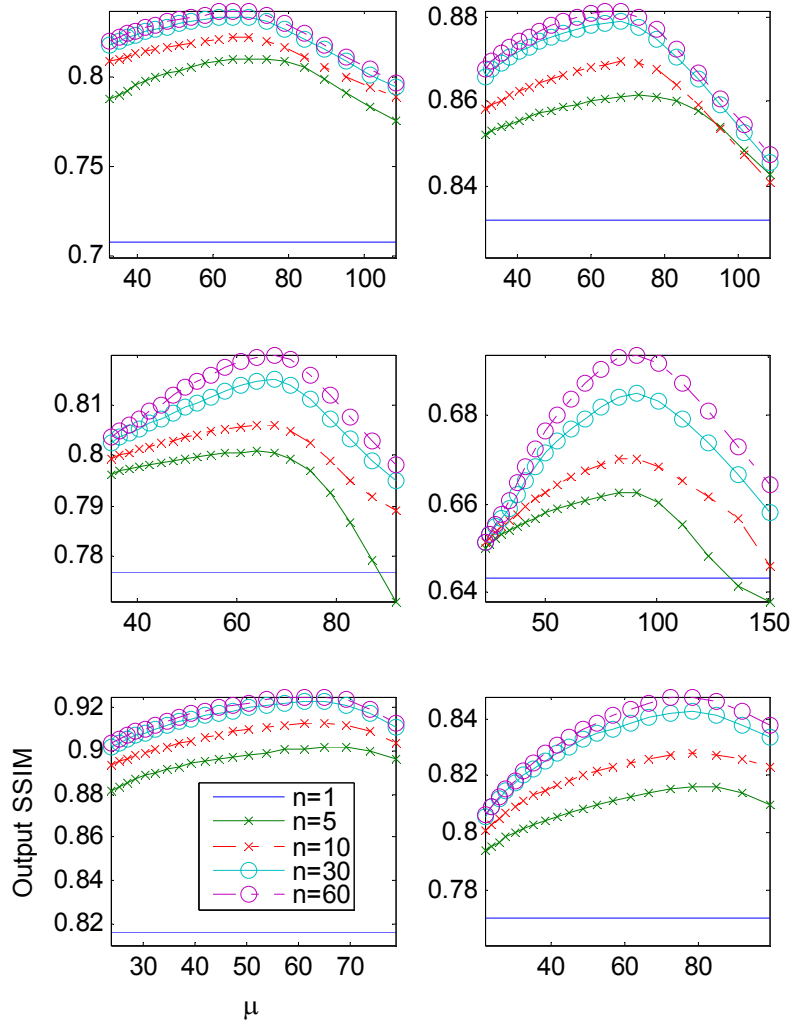

**Figure 2.** Convergence of the output SSIM for various values of step size ($\mu$), for input images *cameraman* (left column) and *lena* (right column), and target histogram of *rice* (top row), linear (middle row), and uniform (bottom row) histograms. *n* is the number of iterations. Improvement over the classic EHS result is visible for all values of $\mu$ shown in the graphs. Note that the value of $\mu$ giving the maximum output SSIM at the $5^{th}$ iteration is also (very close to) optimal for large number of iterations.

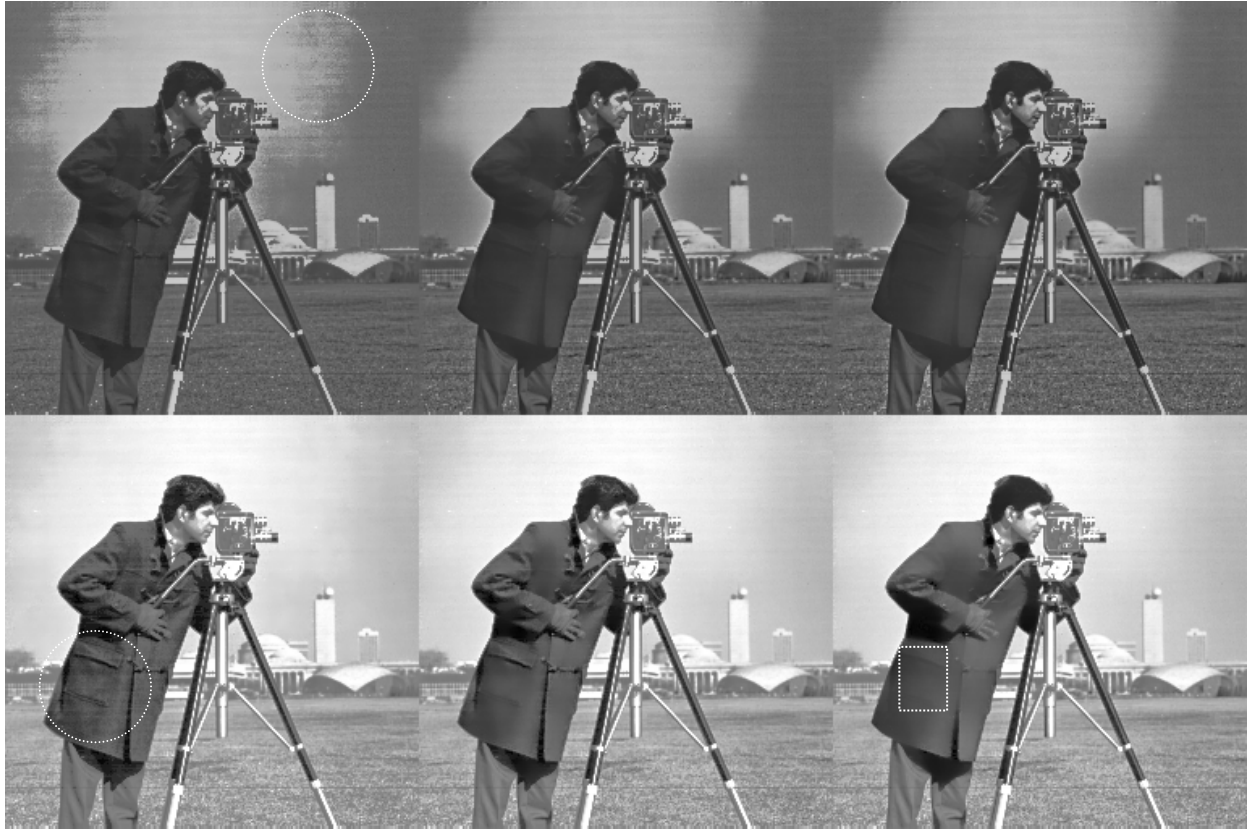

**Figure 3.** Outputs of the proposed method at iterations 2 (left), 12 (middle) and 60 (right), for the target histogram of *rice* (top row), and linear (bottom row). The original *cameraman* is shown in Figure 1. Some areas that received visible improvement are circled. Note that (i) the results of $12^{th}$ and $60^{th}$ iterations do not differ much and (ii) the structure of the area within the dashed rectangle is not very visible in the original *cameraman* either.

### d) Comparison to wavelet-based EHS of [2]

In Table 2, the performance and the runtime of the proposed method (with 12 iterations) in histogram equalization are compared to those of the wavelet-based EHS method of [2] for test images *peppers*, *columbia*, and *plane*. The output images of the two methods are given in Figure 4. In [2], it is shown that Coltuc's method [1] requires about 10% more CPU time and the quality of its outputs is a bit lower (i.e., not visible in some cases), hence we do not repeat Coltuc results for these images. That is while the improvement in the visual quality of the output is visible in all of our experiments (at least 5.5% higher

SSIM index; see bottom row of Figure 3). Note that our code is in Matlab™ and is not optimized for speed.

**Table 2.** Comparison of the proposed method and the method of [2] in terms of output quality (SSIM between the result and the original image in %) and speed (CPU time in seconds), for test images of various sizes.

| Method→ Test image (size)↓ | Wavelet-based EHS of [2] | Proposed (12 iterations) |
|---|---|---|
| *peppers* (512 x 512) | 85.81% in 1.51 sec | 94.94% in 9.89 sec |
| *columbia* (480 x 480) | 76.83% in 1.33 sec | 83.63% in 8.83 sec |
| *plane* (256 x 256) | 53.69% in 0.42 sec | 69.77% in 1.56 sec |

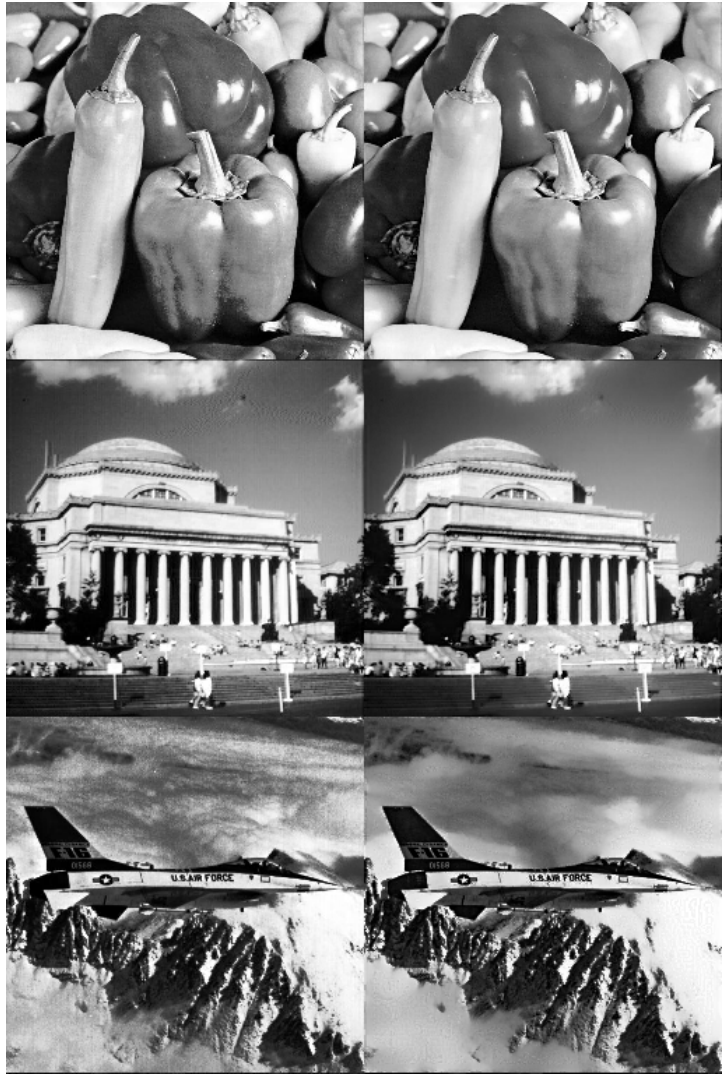

**Figure 4.** Results of histogram equalization using the wavelet-based EHS [2] (left) and the proposed method (right). Better performance of the proposed method is visible in smooth areas in all cases (e.g., better quality of clouds in *plane* is even visible in printed version). SSIM indexes between the results and the originals and the runtimes are listed in Table 2.

## VI. Conclusion

A summary of our contributions follows: (i) The EHS problem is reformulated as an image optimization problem: among all images with the specified histogram, we find the one that is most similar to the input image in terms of SSIM. (ii) A closed-form and computationally simple formula for SSIM gradient is derived by breaking SSIM into linear terms. (iii) The reformulated EHS problem is solved using SSIM gradient ascent. The visual quality of our results is significantly better than that of the existing methods, while keeping the histogram *exactly* as specified. The issues of convergence and computational complexity of the proposed method are addressed. (iv) Since our SSIM-optimum EHS produces visually pleasant results, we show that SSIM is a good image quality metric for the EHS problem.

Failure in strict ordering occurs when two (or more) same-intensity pixels are also the same in terms of the auxiliary information required in step 1 of Algorithm 1. Since [1] and [2] blame distortions in their results on failures in strict ordering, they need to (i) model the input image and the ordering process, (ii) compute a strict ordering failure rate, and (iii) show that the failure rate is under a certain threshold. That is opposed to our method which optimizes the visual fidelity of the result directly and does not need to worry about strict ordering and its failures.

A "minor" advantage of the methods based on strict ordering over the proposed method is that they are "almost" completely reversible [13] (or [1]): Suppose that an EHS based on strict ordering is applied to an image. Assuming that you recorded the histogram of the original image before performing EHS, you can perform the same EHS again, this time with the recorded original histogram as the target, to get back the original image (except for the pixels that failed strict ordering; hence "almost" is used above). For example, we applied [1] to *cameraman* with *rice*'s histogram as the target. Then we applied [1] to the result, with *cameraman*'s histogram as the target. The SSIM between the final result and the original *cameraman* is 0.9964 (i.e., the original is almost exactly recovered by reverse EHS). This figure, if the experiment is performed using our method instead of [1], is 0.9419 when our method is run with 12 iterations. The figure increases to 0.9780, if we allow 150 iterations. Hence the term "minor" is used above.

In [6], an image compression technique is adapted to maximize the minimum of $SSIM_{map}$ (defined in Section II). That is, their goal is to improve the lowest quality (in terms of SSIM) areas of the result, on the premise that the visual attention is attracted to the areas with high distortion. The author disagrees with this philosophy: The visual attention is attracted to the low-quality areas only when the overall quality seems low at the first glance. As an example, consider the kind of puzzle with two similar drawings that one has to find the differences between the two. Since the drawings are very similar (i.e., one has good visual quality, considering the other as the reference), the visual attention is *not* automatically attracted to the differences (i.e., areas with low visual quality). Nevertheless, one may wonder if our proposed scheme can be improved by maximizing the minimum of $\text{SSIM}_{\text{map}}$ rather than maximizing the overall SSIM, and how. Our answer is twofold. First, $\min(\text{SSIM}_{\text{map}}(Y,I))$ is not differentiable, thus it cannot directly replace SSIM in our method. Second, we tried using $X = Y + \mu M \dfrac{\nabla_Y \text{SSIM}(I,Y)}{\text{SSIM}(I,Y)}$ in step 4 of Algorithm 2 to give pixels with lower quality (i.e., smaller $\text{SSIM}(I,Y)$) more improvement. However, we observed that this change deteriorates the visual quality of the final result. This is expected as the *gradient* ascent is supposed to give the fastest growth in SSIM and $\dfrac{\nabla_Y \text{SSIM}(I,Y)}{\text{SSIM}(I,Y)}$ is not SSIM's gradient.

The relative improvement in the visual fidelity of the EHS output in terms of SSIM varies between 5.5 and 18.2 percent in the results reported in Figure 2 for different input images and target histograms. That is because considerable changes in histogram cause distortions in the EHS output that cannot be compensated using the proposed method. For example, linear histogram is very different from the histograms of *lena* and *cameraman*; thus the improvement due to the application of our method in these cases is small (6 and 5.5 percent respectively).

The proposed method cannot be readily used for color images. That is because the quality metric we optimized for (SSIM) works for grayscale images only. However, as suggested in [1], one can apply an EHS to the luminance channel of a color image (I, for example, in HSI color space). In this case, our EHS method can be used to better maintain the visual fidelity of the luminance channel, hence improving the overall quality of the color image.

Derivation of a lower bound and a tighter upper bound on the output improvement in terms of SSIM for given input image and specified target histogram can be a subject of future work. Such bounds are useful for estimation of optimal step size.